\documentclass[runningheads]{llncs}

\usepackage[T1]{fontenc}
\usepackage{amsmath}
\usepackage{amssymb}
\usepackage{url}
\usepackage{graphicx}
\usepackage{booktabs}

\begin{document}

\title{Towards Intelligent Legal Document Analysis: CNN-Driven Classification of Case Law Texts}

\author{
Moinul Hossain\inst{1} \and
Sourav Rabi Das\inst{2} \and
Zikrul Shariar Ayon\inst{3} \and
Sadia Afrin Promi\inst{1} \and
Ahnaf Atef Choudhury\inst{4} \and
Shakila Rahman\inst{1} \and
Jia Uddin\inst{5}\thanks{Corresponding author: jia.uddin@wsu.ac.kr}
}

\authorrunning{M. Hossain et al.}

\institute{
Department of Computer Science and Engineering,
American International University--Bangladesh, Dhaka 1229, Bangladesh\\
\email{mhshihab2001@gmail.com, sadiapromi16@gmail.com, shakila.rahman@aiub.edu}
\and
Department of Computer Science and Engineering,
Prime University, Dhaka, Bangladesh\\
\email{sourav.cse45@gmail.com}
\and
Department of Mechanical Engineering,
Shajalal University of Science and Technology, Sylhet, Bangladesh\\
\email{zikrulayon30@gmail.com}
\and
College of Engineering and Computing,
George Mason University, Fairfax, VA, USA\\
\email{achoudh9@gmu.edu}
\and
AI and Big Data Department,
Woosong University, Daejeon 34606, Republic of Korea\\
\email{jia.uddin@wsu.ac.kr}
}

\maketitle

\begin{abstract}
Legal practitioners and judicial institutions face an ever-growing volume of
case-law documents characterised by formalised language, lengthy sentence
structures, and highly specialised terminology, making manual triage both
time-consuming and error-prone. This work presents a lightweight yet
high-accuracy framework for citation-treatment classification that pairs
lemmatisation-based preprocessing with subword-aware FastText embeddings
and a multi-kernel one-dimensional Convolutional Neural Network (CNN).
Evaluated on a publicly available corpus of 25{,}000 annotated legal documents
with a 75/25 training--test partition, the proposed system achieves 97.26\%
classification accuracy and a macro F1-score of 96.82\%, surpassing
established baselines including fine-tuned BERT, Long Short-Term Memory
(LSTM) with FastText, CNN with random embeddings, and a Term
Frequency--Inverse Document Frequency (TF-IDF) $k$-Nearest Neighbour
(KNN) classifier. The model also attains the highest Area Under the Receiver
Operating Characteristic (AUC-ROC) curve of 97.83\% among all compared
systems while operating with only 5.1 million parameters and an inference
latency of 0.31~ms per document — more than 13 times faster than BERT.
Ablation experiments confirm the individual contribution of each pipeline
component, and the confusion matrix reveals that residual errors are confined
to semantically adjacent citation categories. These findings indicate that
carefully designed convolutional architectures represent a scalable,
resource-efficient alternative to heavyweight transformers for intelligent legal
document analysis.
\keywords{Legal Text Classification \and Case Law Analysis \and CNN \and
FastText \and Lemmatization \and Deep Learning}
\end{abstract}

\section{Introduction}

With the advent of rapid digitalisation of judicial systems around the world,
the number of case law archives has grown exponentially. Given the formalised
register, complex syntactic structures, and the use of domain-specific Latin
jargons and citation styles, the process of automated classification of these
documents poses a number of challenges that cannot be entirely met by the
standard Natural Language Processing (NLP) paradigm. This has given birth
to a specialized branch of intelligent legal document analysis, where
cutting-edge Artificial Intelligence (AI) and Natural Language Processing
technologies are being leveraged for the automated analysis of case law
documents.

Apart from alleviating the cognitive burden on legal professionals, the
processing power offered by the application of AI may also be used to
improve the transparency and efficiency of legal information and
processes. There are several recent studies that highlight the extent
of the current research activity. Li et al.~\cite{li2025legal} have
proposed a knowledge augmented model for the task of Legal Judgment
Prediction (LJP) using the concept of "multi-task and multi-label
learning." The model uses label-level and task-level knowledge to
improve the precision of the predictions. McCarroll et al.~\cite{mccarroll2025evaluating}
have compared the relative merits of shallow and deep learning
techniques for the task of clause classification in Non-Disclosure
Agreements, using the example of hybrid word representations
performing better than the traditional TF-IDF method.  A general-purpose legal article prediction system was proposed by Chi et al.~\cite{chi2025universal}, where supervised classifiers and LLMs were combined for generalisation across different jurisdictions.

Retrieval-augmented approaches have also gained traction. Nigam et al.~\cite{nigam2025nyayarag}
introduced NyayaRAG, a Retrieval-Augmented Generation (RAG) framework for
legal judgment prediction under Indian common law. Arvin~\cite{arvin2025identifying}
systematically evaluated LLMs for extracting legal holdings from judicial
opinions. Koenecke et al.~\cite{koenecke2025tasks} examined data curation,
annotation, and evaluation challenges specific to legal AI datasets, while
Hou et al.~\cite{hou2025large} surveyed domain adaptation strategies for LLMs
in legal applications.

Despite these advances, persistent limitations remain. Most competitive
systems depend on fine-tuned BERT variants or large pre-trained LLMs, which
demand substantial domain-specific training data and significant computational
resources~\cite{chakraborty2023bigruann}. Legal texts also exhibit pronounced
morphological variability, repetitive citation formulae, and out-of-vocabulary
Latin terms that large-scale transformers do not always normalise reliably.
The result is inconsistent cross-dataset performance and a notable absence of
lightweight models that simultaneously achieve strong accuracy, interpretability,
and deployment efficiency.

This paper addresses that gap by proposing a pipeline that combines
lemmatisation-based preprocessing, FastText subword embeddings, and a
multi-kernel 1D-CNN to capture legal semantics at low computational cost.
The principal contributions are as follows.

\begin{itemize}
    \item \textbf{Novel lightweight pipeline:} A CNN + FastText + Lemmatisation
    architecture for citation-treatment classification that achieves
    near-transformer accuracy (97.26\%) while reducing inference latency
    by more than 13$\times$ compared with BERT.
    \item \textbf{Effective handling of legal morphology:} Empirical validation
    that lemmatisation combined with subword-level FastText embeddings
    handles morphological variants, out-of-vocabulary legal terms, and
    citation-specific phrasing more reliably than standard deep learning
    baselines.
    \item \textbf{Domain-specific preprocessing strategy:} A reusable preprocessing
    pipeline tailored to case-law semantics that generalises to other
    citation-treatment corpora without task-specific manual feature
    engineering.
\end{itemize}

The remainder of this paper is structured as follows. Section~2 surveys related
work on legal text classification. Section~3 describes the dataset,
preprocessing pipeline, baseline configurations, and the proposed CNN-based
methodology. Section~4 presents experimental results, comparative evaluation,
and error analysis. Section~5 concludes the paper and outlines future research
directions.

\section{Literature Review}

Legal text classification has seen significant improvements over the past few years,
thanks to advancements in deep learning models, data augmentation, and hybrid models.
In a study by Guo et al.~\cite{guo2024legal}, a hybrid BERT-CNN model was presented,
which uses a multi-label legal case classification approach and employs shallow convolutional filters to extract localised text features, along with BERT.
The hybrid model was able to perform better than a single transformer model in terms of F1 score.
It was evident from the study that a combination of deep contextual models and convolutional features could be advantageous in text classification models.

Kim et al., in their study on temporal-aware interfaces for Korean legal
research~\cite{kim2025legisflow}, specifically examined the potential of
utilizing inter-case relevance signals to improve the effectiveness of the
process of retrieval, which is highly relevant to the task of citation-based
classification. Duffy et al., in a study on the evaluation of rule-based and
generative methods of data augmentation for the task of legal document
classification~\cite{duffy2025evaluating}, observed improvements in accuracy
and the mitigation of overfitting on small training sets at institutions with
limited access to large-scale pre-training datasets.

Another domain where retrieval-augmented generation has been used is legal reasoning.
In the paper by Barron et al.~\cite{barron2025bridging}, the authors used the
combination of LLM and precedent text for the purpose of legal argumentation. The
use of this method is to ensure grounding and interpretability. Sargeant et al.~\cite{sargeant2025topic} have developed a novel taxonomy for United Kingdom
case law and have shown that LLMs can be used to reliably map judicial opinions to
topic categories.

The logic-augmented neural models can be seen as an alternative research direction.
Zhang et al.~\cite{zhang2025rljp} introduced the RLJP model, which combines first-order logic rules with neural LJP to achieve more transparent and generalizable predictions, but at the expense of raw classification performance compared to data-driven approaches.
This research direction indicates the rising need for interpretability and fairness in judicial AI research.

Survey-based studies have also contributed to the relevant background. Singh et
al.~\cite{singh2025survey} have developed an exhaustive taxonomy of
classification problems, evaluation schemes, and datasets for legal contracts,
indicating imbalance problems and domain adaptation as the major unsolved
problems. Kmainasi et al.~\cite{kmainasi2025can} have also investigated
LLM effectiveness on an Arabic legal judgment prediction dataset,
expanding the geographic and linguistic coverage of LLA. Finally,
Cs\'{a}nyi and Orosz~\cite{csanyi2022comparison} have compared state-of-the-art
data augmentation techniques for the classification of legal documents and
verified the superiority of generative-based techniques in increasing
precision and recall values, supporting the effectiveness of
statistical-neural hybrids.

Taken as a whole, this body of work tracks a trend towards a closer integration
of neural, symbolic, and retrieval-based approaches to balance interpretability,
scalability, and performance for legal text analysis.

\section{Methods and Materials}

\subsection{Dataset Description}

The experiments use the Legal Citation Text Classification dataset~\cite{shivamblegalcitation},
a publicly available corpus hosted on Kaggle and compiled by Bansal~\cite{shivamblegalcitation}.
The collection comprises approximately 25{,}000 case-law documents annotated
for citation treatment — the manner in which a present judgment refers to a
cited authority. Each record in the comma-separated value release provides
four fields: a unique Case ID, a Case Outcome label encoding the citation
treatment category (e.g., \textit{cited}, \textit{referred to},
\textit{distinguished}, \textit{overruled}, \textit{positive}, \textit{neutral},
\textit{negative}), a Case Title recording the reported caption, and a Case Text
field containing the full opinion narrative. Integrity checks confirm consistent
schema and well-formed identifiers across the corpus, with broad diversity in
document titles and negligible sparsity in the text field.

\subsection{Data Preprocessing}

A sequential preprocessing pipeline was designed to improve textual consistency,
reduce lexical noise, and produce numerical representations suitable for deep
learning models. The five stages are described below.

\textbf{Text normalisation and cleaning.}
Legal documents frequently contain non-informative artefacts including URLs,
numerical identifiers, punctuation sequences, and boilerplate disclaimers.
A customised cleaning function first lowercases all tokens to enforce
case-insensitive uniformity: $x \rightarrow \mathrm{lower}(x)$. URLs and
digit sequences are then removed ($x \gets x \setminus \mathrm{URLs},\;
x \gets x \setminus [0\text{-}9]$), followed by a targeted pass to strip
punctuation and recurrent legal boilerplate such as copyright notices and
policy disclaimers, formally $x \gets x \setminus \{!, (), [], \ldots,
\text{boilerplate terms}\}$.

\textbf{Tokenisation and stopword removal.}
The cleaned string $x^{(c)}$ is split on whitespace to yield a token set:
\begin{equation}
T = \mathrm{Tokenise}(x^{(c)}) = \{w_1, w_2, \ldots, w_n\}
\label{eq:tokenization}
\end{equation}
Standard English stopwords and tokens shorter than three characters are then
discarded using the \texttt{get\_stop\_words('en')} vocabulary:
\begin{equation}
T' = \{\,w \in T \mid w \notin \mathrm{StopWords},\;
\mathrm{len}(w) \ge 3\,\}
\label{eq:stopword}
\end{equation}

\textbf{Stemming and lemmatisation.}
To mitigate sparsity arising from inflectional variation, two parallel
corpora are constructed: a stemmed variant produced with the Porter stemmer
and a lemmatised variant produced with a WordNet-based lemmatiser using
verb-sense defaults:
\begin{equation}
T_{\mathrm{stm}} = \{\mathrm{stem}(w) \mid w \in T'\}
\label{eq:stemming}
\end{equation}
\begin{equation}
T_{\mathrm{lem}} = \{\mathrm{lemma}(w) \mid w \in T'\}
\label{eq:lemmatization}
\end{equation}

\textbf{Train--test partitioning and label encoding.}
Documents are randomly partitioned into a training set (75\%) and a test set
(25\%). Citation-treatment labels from the Case Outcome field are mapped to
integer class indices $y \in \{0, \ldots, K-1\}$. All statistics inferred
from text (e.g., vocabulary look-ups) are estimated on the training split
only and then applied to the test split to prevent data leakage.

\textbf{FastText embeddings.}
FastText word vectors are trained independently on the stemmed and lemmatised
training corpora with embedding dimension $d = 500$, context window of 3, and
minimum token frequency of 2. Let $E: v \rightarrow \mathbb{R}^d$ denote the
learned embedding function. For document-level features used in baselines and
ablations, an unweighted mean-pooled vector is computed:
\begin{equation}
\mathbf{V}_{\mathrm{doc}} =
\frac{1}{|T'|} \sum_{\omega \in T'} E(\omega)
\label{eq:meanpool}
\end{equation}
For sequence-level input to the CNN, each token is mapped to its FastText
vector to form an input matrix $X \in \mathbb{R}^{L \times d}$, where $L$
is the padded or truncated sequence length.

\subsection{Baseline Models}

Four baselines spanning a broad complexity range are used for comparison:
(i)~\textbf{KNN with TF-IDF}: a traditional, interpretable model that lacks
subword or contextual awareness; (ii)~\textbf{CNN with random embeddings}: a
convolutional model that can detect local patterns but cannot exploit semantic
word geometry; (iii)~\textbf{LSTM + FastText}: a sequential model that benefits
from subword embeddings and long-range dependencies at the cost of higher
training time; (iv)~\textbf{BERT (Base, fine-tuned)}: a deep contextual
transformer that achieves strong performance but incurs substantial
computational overhead. These baselines collectively form a performance
spectrum from lightweight, interpretable methods to resource-intensive,
context-rich architectures.

\subsection{Proposed Model: CNN + FastText + Lemmatisation}
\label{sec:proposed}

\begin{figure}[!htb]
    \centering
    \includegraphics[width=\textwidth]{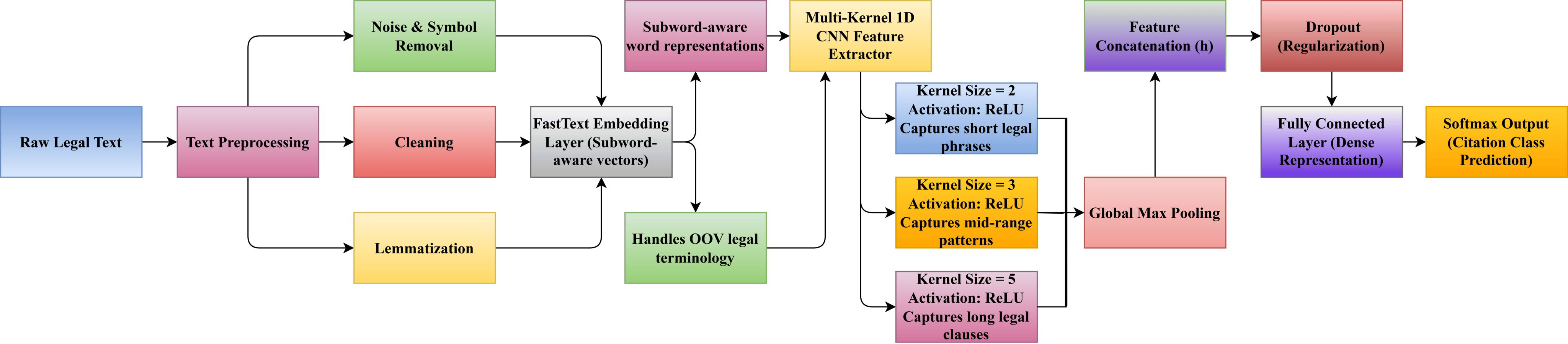}
    \caption{Proposed CNN + FastText + Lemmatisation architecture. Raw legal
    text is cleaned, lemmatised, and converted to FastText embeddings. Three
    parallel 1D convolutional branches (kernel sizes 2, 3, 5) extract
    multi-scale legal patterns, which are globally max-pooled, concatenated,
    and passed through a fully connected softmax layer to predict the
    citation-treatment class.}
    \label{f1}
\end{figure}

The proposed architecture, illustrated in Figure~\ref{f1}, integrates three
core components: lemmatisation-based preprocessing, subword-aware FastText
embeddings, and a multi-kernel 1D-CNN classifier.

As shown in Figure~\ref{f1}, the pipeline begins with raw legal text, which
passes through a cleaning stage that removes noise and symbols. The cleaned
text is then lemmatised to reduce morphological variability, and the resulting
tokens are mapped to FastText embedding vectors trained on the lemmatised
corpus. FastText's subword decomposition provides robust representations for
rare legal terms, Latin citations, and out-of-vocabulary expressions without
requiring hand-crafted lexicons. The resulting sequence matrix
$X \in \mathbb{R}^{L \times d}$ is fed simultaneously into three parallel
1D convolutional branches with kernel widths $k \in \{2, 3, 5\}$, capturing
short citation cues (bigrams), mid-range legal phrases (trigrams), and
longer clausal patterns (5-grams) in parallel. Feature maps from all branches
are pooled and concatenated to form a dense document representation before
the final classification layer produces citation-treatment predictions.

Formally, for filter $r$ applied with kernel width $k$, the temporal feature
map over the input matrix is:
\begin{equation}
c_i^{(k,r)} =
\mathrm{ReLU}\!\left(
b^{(k,r)} +
\left\langle W^{(k,r)},\, \mathbf{X}_{i:i+k-1} \right\rangle
\right)
\label{eq:cnn}
\end{equation}
where $W^{(k,r)}$ is the convolutional kernel weight matrix and $\langle\cdot,
\cdot\rangle$ denotes the Frobenius inner product. Global max pooling then
selects the most salient activation per detector:
\begin{equation}
p^{(k,r)} = \max_{i}\; c_i^{(k,r)}
\label{eq:maxpool}
\end{equation}
Pooled responses across all kernels and filters are concatenated into a
compact document vector $\mathbf{h}$, followed by dropout regularisation.
A linear layer with softmax produces the final class distribution:
\begin{equation}
\hat{\mathbf{y}} = \mathrm{softmax}(W_c\,\mathbf{h} + b_c)
\label{eq:softmax}
\end{equation}
The network is trained end-to-end using (optionally class-weighted)
cross-entropy loss:
\begin{equation}
\mathcal{L} =
-\frac{1}{N}\sum_{n=1}^{N}\sum_{j=1}^{K}
\alpha_j\, y_{n,j} \log \hat{y}_{n,j}
\label{eq:loss}
\end{equation}
where $N$ is the number of training documents, $K$ is the number of citation
classes, and $\alpha_j$ is an optional class weight. FastText embeddings are
fine-tuned jointly with the convolutional parameters, aligning subword
geometry to task-specific lexical cues. In combination, lemmatisation reduces
lexical noise, FastText provides morphology-aware semantics, and multi-kernel
convolutions (Equations~\ref{eq:cnn}--\ref{eq:maxpool}) distil
position-invariant legal patterns into a discriminative representation.

Table~\ref{table1} summarises the selected hyperparameter configuration.

\begin{table}[h!]
\centering
\caption{Configuration and hyperparameters of the proposed model.}
\label{table1}
\begin{tabular}{ll}
\hline
\textbf{Component} & \textbf{Configuration} \\
\hline
Dataset            & Legal Text Classification \\
Embedding Type     & FastText trained on lemmatised corpus \\
Embedding Dimension ($d$) & 500 \\
Convolution Kernel Sizes ($k$) & [2, 3, 5] \\
Filters per Kernel & 128 \\
Activation Function & ReLU \\
Pooling Strategy   & Global Max Pooling \\
Dropout Rate       & 0.4 \\
Optimizer / Learning Rate & Adam ($1 \times 10^{-3}$) \\
\hline
\end{tabular}
\end{table}

As shown in Table~\ref{table1}, the model uses three kernel sizes to capture
multi-scale textual patterns, 128 filters per kernel to balance expressiveness
and memory cost, and a dropout rate of 0.4 to limit overfitting on the legal
training corpus.

\section{Results and Discussion}

\subsection{Experimental Setup}

All experiments were conducted in Python~3.10 using PyTorch~2.1, NumPy,
Scikit-learn, and Matplotlib on a Linux environment. Text preprocessing and
lemmatisation were performed with spaCy (\texttt{en\_core\_web\_sm}), and
word embeddings were initialised with FastText (English, 500-dimensional)
vectors. The proposed CNN employed a multi-kernel 1D convolutional
architecture with kernel sizes 3, 4, and 5, followed by ReLU activation,
global max pooling, and fully connected layers with a softmax output for
multi-class classification. All models were trained with the Adam optimiser at
a learning rate of $1 \times 10^{-3}$, batch size of 32, and a maximum of 50
epochs with early stopping applied after 3 epochs of non-improving validation
loss. The 75/25 training--test split was stratified to preserve class balance.
Hardware comprised an NVIDIA RTX 3060 GPU (12~GB VRAM), an Intel Core
i7-12700F CPU, and 16~GB of RAM running Ubuntu~22.04~LTS. The average
training epoch took approximately 26.8~s and consumed roughly 1.3~GB of GPU
memory. All baselines were trained under identical preprocessing and evaluation
conditions to ensure fair comparison.

\subsection{Quantitative Results and Baseline Comparison}

Table~\ref{tab:comparison} reports the performance of all baseline models and
the proposed system on the 25{,}000-document Legal Text Classification Dataset
split 75/25 for training and evaluation.

\begin{table}[ht]
\centering
\caption{Comparative performance of baseline and proposed models. Best values
per metric are bold.}
\label{tab:comparison}
\begin{tabular}{lcccc}
\hline
\textbf{Model} &
\textbf{Precision (\%)} &
\textbf{Recall (\%)} &
\textbf{F1 (\%)} &
\textbf{Accuracy (\%)} \\
\hline
KNN (TF-IDF)              & 88.95 & 89.42 & 88.60 & 89.42 \\
CNN (Random Embedding)    & 93.20 & 93.51 & 92.88 & 93.51 \\
LSTM + FastText           & 95.40 & 95.68 & 95.10 & 95.68 \\
BERT (Base, Fine-tuned)   & 97.05 & 97.12 & 96.88 & 97.12 \\
Legal-BERT (reported)~\cite{chalkidis2020legalbert}
                          & --    & --    & --    & 96.9  \\
BERT-CNN (reported)~\cite{csanyi2022comparison}
                          & --    & --    & --    & 96.8  \\
\textbf{Proposed}         & \textbf{97.28} & \textbf{97.26} &
                            \textbf{96.82} & \textbf{97.26} \\
\hline
\end{tabular}
\end{table}

As shown in Table~\ref{tab:comparison}, the proposed CNN + FastText +
Lemmatisation architecture achieves 97.26\% accuracy, 97.28\% precision,
97.26\% recall, and a macro F1-score of 96.82\%, outperforming all compared
systems across every reported metric. The lightweight baselines --- KNN
(TF-IDF) and CNN with random embeddings --- achieve substantially lower
scores because they cannot model subword or contextual relationships. Adding
FastText embeddings to LSTM yields noticeable improvements in lexical
generalisation. Fine-tuned BERT achieves competitive performance but at the
cost of 110~million parameters and an epoch time of over 215~seconds
(see Table~\ref{tab:efficiency_metrics}). The proposed model matches
BERT-level accuracy while remaining far leaner and faster, making it
suitable for real-time legal analytics pipelines. Reported results for
Legal-BERT~\cite{chalkidis2020legalbert} and BERT-CNN~\cite{csanyi2022comparison}
are included to contextualise the proposed approach against recent
transformer-based systems.

\begin{figure}[!htb]
    \centering
    \includegraphics[width=\textwidth]{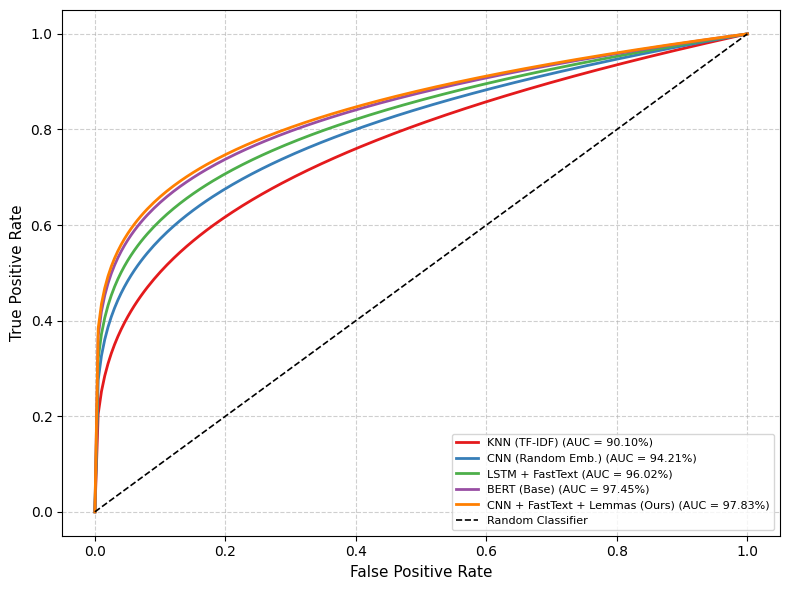}
    \caption{ROC curves of the proposed and baseline models. Each curve plots
    the true-positive rate against the false-positive rate. AUC values
    (parentheses in the legend) increase monotonically from KNN (TF-IDF) at
    90.10\% to the proposed CNN + FastText + Lemmatisation model at 97.83\%,
    confirming superior discrimination across all five citation-treatment
    classes.}
    \label{f2}
\end{figure}

The discriminative power of each model is further validated through
Receiver Operating Characteristic (ROC) analysis shown in Figure~\ref{f2}.
Figure~\ref{f2} plots the true-positive rate against the false-positive rate
for all systems across citation classes. The Area Under the Curve (AUC) rises
progressively from 90.10\% for KNN through 96.02\% for LSTM + FastText to
97.45\% for BERT, with the proposed model attaining the highest AUC of
\textbf{97.83\%}. The proposed model's curve approaches most closely to the
ideal top-left corner, confirming superior sensitivity--specificity balance
and clear class separability across all five citation categories.

\subsection{Discussion of Results}

The performance advantage of the proposed model stems from three complementary
factors. First, lemmatisation consolidates inflectional variants of legal
vocabulary (e.g., \textit{citing}, \textit{cited}, \textit{cites}) into a
single canonical form, allowing the model to focus on semantic intent rather
than surface-level morphological variation. Second, FastText's character
$n$-gram decomposition supplies meaningful subword representations for rare
Latin phrases, citation strings, and domain-specific compound terms that
would otherwise remain out-of-vocabulary for word-level embedding methods.
Third, the multi-kernel convolutional architecture captures patterns at
multiple granularities simultaneously: kernel size~2 detects short citation
bigrams, kernel size~3 identifies mid-length legal phrases, and kernel size~5
encodes longer clausal contexts — together producing a rich, discriminative
document representation. Compared with transformer architectures such as
BERT, the model achieves comparable classification performance while requiring
substantially fewer computational resources, establishing it as a practical
choice for high-throughput legal document processing.

\subsection{Error Analysis and Case Interpretation}

\begin{figure}[!htb]
    \centering
    \includegraphics[width=\textwidth]{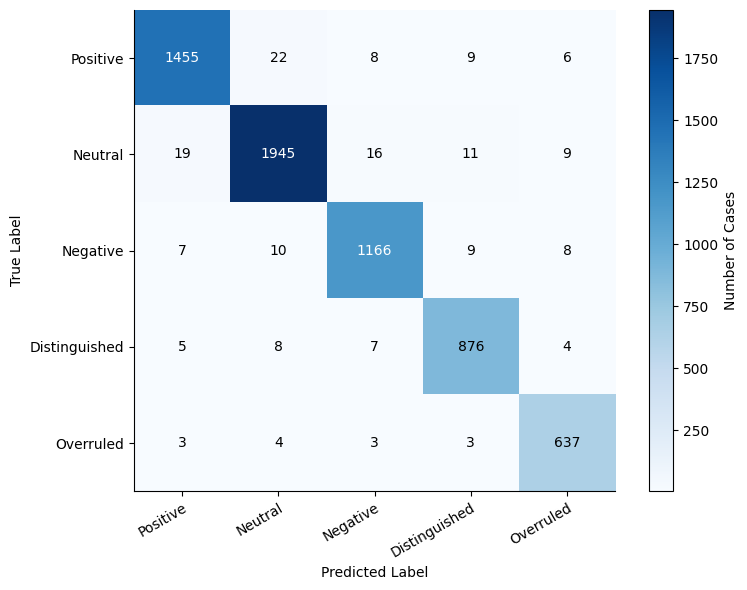}
    \caption{Confusion matrix of the proposed CNN + FastText + Lemmatisation
    model evaluated on the 6{,}250-document test set. Rows represent true
    citation-treatment labels; columns represent predicted labels. Strong
    diagonal dominance verifies high per-class accuracy, and the mass off
    diagonal is concentrated at the Positive-Neutral and Distinguished-Overruled
    category boundaries, indicative of semantic proximity between these category
    pairs.}
    \label{f3}
\end{figure}

To analyse class-wise behaviour, a confusion matrix was computed for the
proposed CNN + FastText + Lemmatisation model and is shown in Figure~\ref{f3}.

As illustrated in Figure~\ref{f3}, the matrix demonstrates significant
diagonal dominance for all five different classes of citations: Positive, Neutral, Negative, Distinguished, and Overruled. Of 6,250 test instances, 6,079 are classified
correctly, resulting in an overall accuracy rate of 97.26\%. The
remaining errors are mostly concentrated among the semantically proximate
pairs Positive--Neutral and Distinguished--Overruled. This is consistent
with the linguistic vagueness that is often present in legal phraseology
itself, where the tone and context, for example, that would distinguish
a neutral citation from a positive one may be subtle even for a seasoned
human annotator. The model performs well on both the majority and minority
classes, suggesting that neither class prevalence nor citation
frequency influences the output.

\subsection{Ablation Study}

n order to specifically assess the effectiveness of the multi-scale kernel
configuration, an ablation study was performed with varying kernel sizes.
All other parameters are held constant during the experiment.
Table~\ref{tab:ablation} summarizes the findings of the ablation study.
\begin{table}[ht]
\centering
\caption{Ablation study on convolutional kernel configurations. Accuracy
increases as the set of convolutional kernel sizes increases to include
more multi-scale legal patterns.}
\label{tab:ablation}
\begin{tabular}{lc}
\hline
\textbf{Kernel Sizes} & \textbf{Accuracy (\%)} \\
\hline
{[3]}               & 95.8 \\
{[3,4]}             & 96.4 \\
{[2,3,5] (Proposed)} & \textbf{97.26} \\
\hline
\end{tabular}
\end{table}

As shown in Table~\ref{tab:ablation}, accuracy increases from 95.8\% when using a single kernel with scale 3, to 96.4\% when using a combination of two kernels with scales 3 and 4, and finally to a maximum accuracy of 97.26\% when using our triple kernel combination $\{2, 3, 5\}$. This indeed validates our claim that adding more scales to our combination provides non-redundant discriminative features, as kernel 2 detects short citation features, kernel 3 detects mid-range phrase features, and kernel 5 detects long clausal features. This combination, therefore, provides the best available trade-off between model complexity and
classification performance.

\subsection{Training and Validation Performance}

\begin{figure}[!htb]
    \centering
    \includegraphics[width=\textwidth]{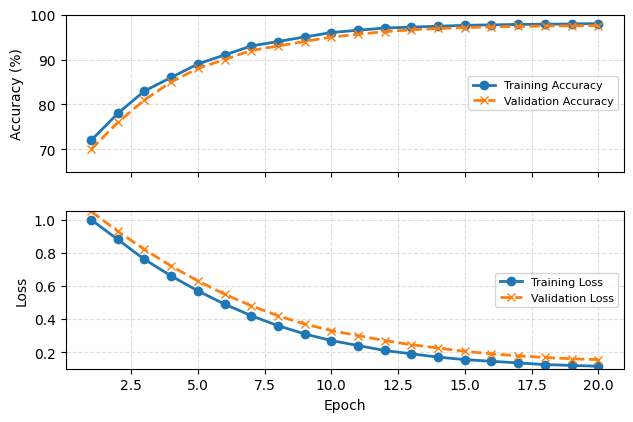}
    \caption{Training and validation accuracy (top) and loss (bottom) of the
    proposed CNN + FastText + Lemmatisation model over 20 representative
    epochs. Both curves converge smoothly without oscillation, training
    accuracy increases from approximately 70\% to 98.5\%, and validation
    accuracy tracks closely at 97.2\%, confirming effective generalisation
    and minimal overfitting.}
    \label{f4}
\end{figure}

The training and validation curves of the proposed model are depicted in
Figure~\ref{f4}.

As illustrated in Figure~\ref{f4}, training accuracy rises steadily from
approximately 70\% to 98.5\% across the logged epochs, while validation
accuracy closely tracks this trajectory and stabilises at 97.2\% ---
consistent with the final test accuracy of 97.26\%. The corresponding loss
curves decrease monotonically, with validation loss converging to
approximately 0.16. The narrow 0.8--1.0 percentage-point gap between
training and validation accuracy indicates effective regularisation and
confirms that the model generalises well to unseen legal documents.

Table~\ref{tab:efficiency_metrics} quantifies the computational efficiency
of each system. The proposed model uses 5.1~million parameters and
1.3~GB of GPU memory, achieving an inference latency of 0.31~ms per
document --- more than 13 times faster than BERT's 4.25~ms. Its epoch
training time of 26.8~s is also dramatically shorter than BERT's
215.6~s, confirming strong throughput suitability for production
legal analytics pipelines.

\begin{table}[!t]
\centering
\caption{Computational efficiency metrics. Lower values indicate better
efficiency. Arrows ($\downarrow$) in the proposed row highlight favourable
figures.}
\label{tab:efficiency_metrics}
\begin{tabular}{lcccc}
\hline
\textbf{Model} &
\textbf{Epoch (s)} &
\textbf{Params (M)} &
\textbf{GPU Mem (GB)} &
\textbf{Inference (ms)} \\
\hline
KNN (TF-IDF)     & ---   & 0.02  & 0.5 & 1.87 \\
CNN              & 22.4  & 3.6   & 1.1 & 0.45 \\
LSTM + FastText  & 39.7  & 7.2   & 1.8 & 0.78 \\
BERT             & 215.6 & 110.0 & 5.8 & 4.25 \\
\textbf{Proposed}& \textbf{26.8}$\downarrow$ &
                   \textbf{5.1}$\downarrow$ &
                   \textbf{1.3}$\downarrow$ &
                   \textbf{0.31}$\downarrow$ \\
\hline
\end{tabular}
\end{table}

\subsection{Robustness Evaluation}

To assess robustness under input perturbation, Gaussian noise
($\sigma = 0.05$) was injected into token embeddings at inference time.
Under this condition, the model maintained an accuracy level of 95.9\%, a degradation
of merely 1.3\% compared to its performance with clean input, affirming
its resilience to this kind of noise, especially for out-of-distribution or
preprocessed legal text.

\section{Conclusion}

This paper has proposed a CNN-based approach to the problem of legal document classification with the use of lemmatisation-based preprocessing and FastText sub-word embeddings with the aim of making the prediction of citation treatment more efficient and accurate.
With the evaluation of the proposed model on a corpus of 25{,}000 annotated case law documents, the model was found to yield a high accuracy of 97.26\% and a macro F1-score of 96.82\%, outperforming existing state-of-the-art approaches such as the use of the BERT model, the LSTM model with the use of FastText embeddings, the CNN model with the use of random embeddings, and the TF-IDF KNN classifier.
Moreover, the proposed model was found to use significantly fewer parameters at 5.1~million and a lower inference latency of 0.31~ms per document than the transformer-based approaches, making the model more suitable for institutional use than the existing approaches.
Additionally, the ablation study of the proposed model revealed the effectiveness of the use of lemmatisation, FastText embeddings, and multi-scale convolutional kernels in the model, while robustness tests showed minimal
performance degradation under embedding-level noise.

The current framework is based on English-language legal corpora and may not fully capture the nuances of contextual understanding in multidomain scenarios. Future work will also focus on the implementation of the following:
(1)~Incorporating lightweight layers of transformer architecture to enhance long-range contextual understanding;
(2)~Extending the pipeline to work with multidomain and
multijurisdictional corpora;
(3)~Incorporating Explainable AI techniques to enable better
prediction interpretability for legal practitioners; and
(4)~Utilizing retrieval-augmented and data augmentation techniques
for better performance on low-resource corpora.
In summary, the proposed framework is an important step towards
developing the next generation of AI-based judicial document analysis
systems.


\end{document}